\documentclass[conference]{IEEEtran}
\IEEEoverridecommandlockouts
\usepackage{cite}
\usepackage{amsmath,amssymb,amsfonts}
\usepackage{algorithmic}
\usepackage{graphicx}
\usepackage{textcomp}
\usepackage{xcolor}

\usepackage{multirow}
\usepackage{booktabs} 
\usepackage[colorlinks=true, urlcolor=blue, linkcolor=red]{hyperref}

\def\BibTeX{{\rm B\kern-.05em{\sc i\kern-.025em b}\kern-.08em
    T\kern-.1667em\lower.7ex\hbox{E}\kern-.125emX}}
\begin{document}

\title{Adapting a Segmentation Foundation Model for \\Medical Image Classification
}

\author{
\IEEEauthorblockN{
   Pengfei Gu\IEEEauthorrefmark{1},
   Haoteng Tang\IEEEauthorrefmark{1},
   Islam A. Ebeid\IEEEauthorrefmark{2},
   Jose A. Nunez\IEEEauthorrefmark{1},
   Fabian Vazquez\IEEEauthorrefmark{1},
   Diego Adame\IEEEauthorrefmark{1},\\
   Marcus Zhan\IEEEauthorrefmark{3},
   Huimin Li\IEEEauthorrefmark{4},
   Bin Fu\IEEEauthorrefmark{1},
   Danny Z.\ Chen\IEEEauthorrefmark{5}
}
\IEEEauthorblockA{\IEEEauthorrefmark{1}Department of Computer Science, University of Texas Rio Grande Valley, Edinburg, TX 78539, USA}
\IEEEauthorblockA{\IEEEauthorrefmark{2}Department of Computer Science, Texas Woman's University, Denton, TX 76204, USA}
\IEEEauthorblockA{\IEEEauthorrefmark{3}Sewickley Academy, Sewickley, PA 15143, USA}
\IEEEauthorblockA{\IEEEauthorrefmark{4}Department of
Mathematical Sciences, The University of Texas at Dallas, Richardson, TX
75080, USA}
\IEEEauthorblockA{\IEEEauthorrefmark{5}Department of Computer Science and Engineering, University of Notre Dame, Notre Dame, IN 46556, USA}
}


\maketitle

\begin{abstract}
Recent advancements in foundation models, such as the Segment Anything Model (SAM), have shown strong performance in various vision tasks, particularly image segmentation, due to their impressive zero-shot segmentation capabilities.
However, effectively adapting such models for medical image classification is still a less explored topic.
In this paper, we introduce a new framework to adapt SAM for medical image classification.
First, we utilize the SAM image encoder as a feature extractor to capture segmentation-based features that convey important spatial and contextual details of the image, while freezing its weights to avoid unnecessary overhead during training.
Next, we propose a novel Spatially Localized Channel Attention (SLCA) mechanism to compute spatially localized attention weights for the feature maps.
The features extracted from SAM’s image encoder are processed through SLCA to compute attention weights, which are then integrated into deep learning classification models to enhance their focus on spatially relevant or meaningful regions of the image, thus improving classification performance.
Experimental results on three public medical image classification datasets demonstrate the effectiveness and data-efficiency of our approach.
\end{abstract}

\begin{IEEEkeywords}
Segmentation foundation model, deep learning, medical image classification
\end{IEEEkeywords}

\section{Introduction} \label{intro}
%
Medical image classification is a critical problem in computer-aided diagnosis, requiring high accuracy and robustness.
While deep learning (DL) models, such as ResNet~\cite{he2016deep}, SENet~\cite{hu2018squeeze}, and Swin Transformer v2~\cite{liu2022swin}, have achieved remarkable success in this domain, they often face several challenges:
(1) Capturing spatially localized features crucial for distinguishing subtle anatomical variations;
(2) neglecting valuable segmentation-based information that may provide further insights into the spatial structures of medical images.

The Segment Anything Model (SAM)~\cite{kirillov2023segment}, a foundation model originally designed for natural image segmentation tasks, offers the potential to bridge this gap with its powerful image encoder that excels at capturing rich features from diverse visual patterns.
A number of recent studies have explored customizing SAM for medical image segmentation, by either directly applying SAM~\cite{ deng2023segment,mattjie2023exploring,huang2023segment,zhang2023input,zhang2023samdsk} or fine-tuning SAM for medical images 
\cite{ma2023segment,li2023polyp,zhang2023customized}.
However, SAM has yet to be fully adapted for medical image classification.
SAMAug-C~\cite{gu2024boosting} proposed a new augmentation method that leverages SAM for augmenting raw image inputs to improve medical image classification. However, this method did not fully explore SAM's capability to capture spatially localized features that provide valuable segmentation-based information.

To address this challenge, we propose a new framework to effectively adapt SAM for medical image classification.
This work addresses two pivotal questions regarding medical image classification:
\textbf{(I) How to efficiently utilize SAM to extract segmentation-based features?
(II) How to effectively integrate such features into DL classification models to enhance their performance?}

For question (I), we propose to use SAM's image encoder as a feature extractor to capture spatially localized features enriched with segmentation-based information.
Specifically, we extract feature maps from the patch embedding, the first, middle, and last blocks of the Transformer blocks, and the Convolution (Conv) block.
These extracted feature maps are then incorporated into different stages of the DL classification models.
The weights of the SAM image encoder are frozen during model training, enabling efficient adaptation.

For question (II), we introduce a novel attention mechanism called Spatially Localized Channel Attention (SLCA).
SLCA is applied to compute the attention weights of the captured features from the SAM image encoder, which are then used to enable the intermediate features of the DL classification models to focus on what is locally (spatially) meaningful.
This attention mechanism helps effective incorporation of the extracted features into the DL classification models, guiding them to focus on spatially meaningful regions and leading to more accurate and interpretable predictions.
\begin{figure*}[ht!]
    \centering
    \includegraphics[width=1.0\textwidth]{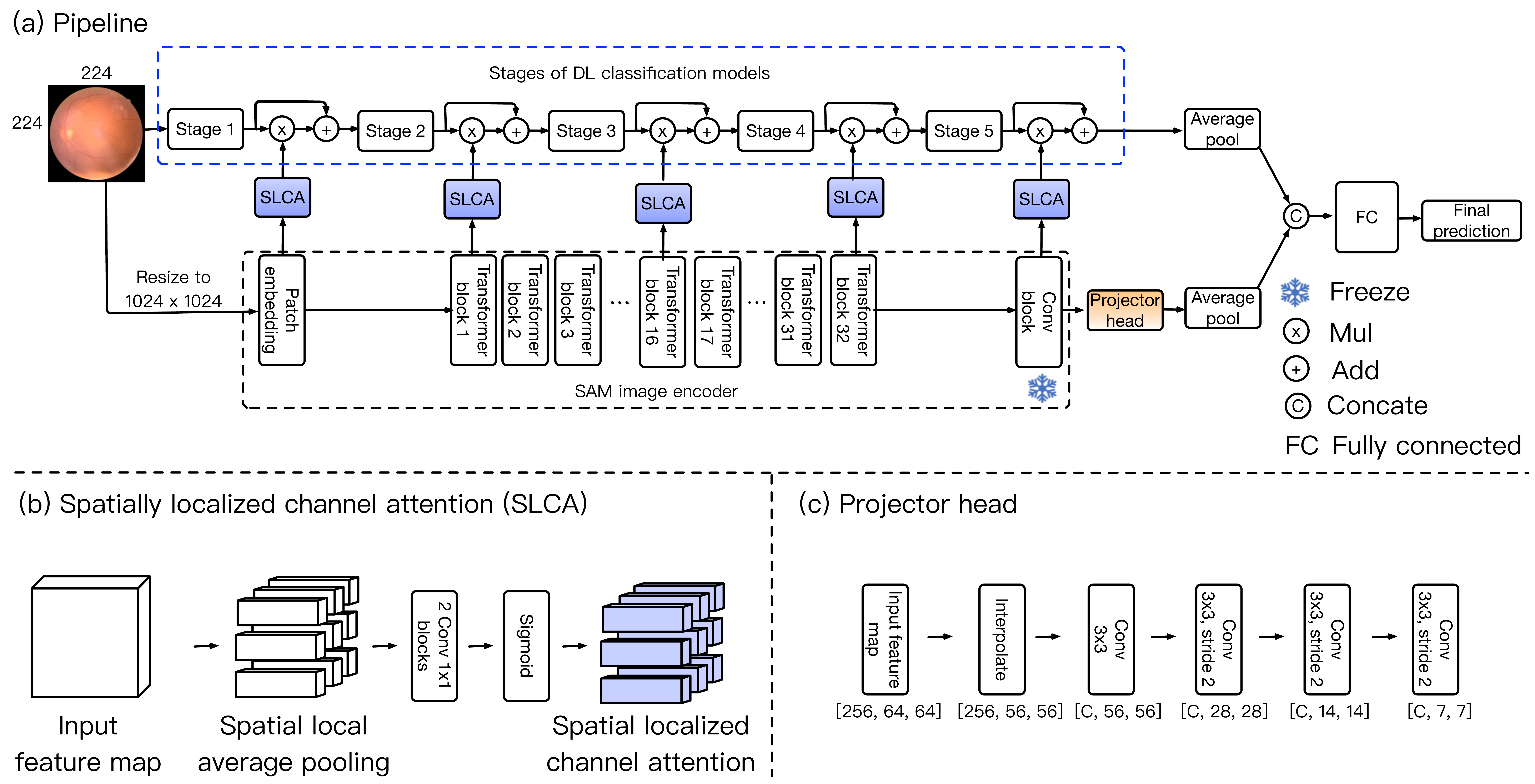}
   \caption{An overview of our framework. In $[C, H, W]$, $C$ represents the number of channels, and $H$ and $W$ represent the height and width of a feature map, respectively. For simplicity, the BN and ReLU that follow each Conv layer are not shown.}
    \label{fig:framework}
\end{figure*}

In summary, our main contributions are as follows: (1) We efficiently capture segmentation-based features using the SAM image encoder as a frozen feature extractor.
(2) We introduce the SLCA mechanism to compute spatially localized attention weights, effectively incorporating the extracted segmentation-based features into DL classification models, enhancing their focus on spatially relevant regions in input images and improving their performance.
(3) We demonstrate the effectiveness and data efficiency of our new approach with extensive experiments on three public medical image datasets.

\section{Methodology} \label{method}
Fig.~\ref{fig:framework}(a) shows the pipeline of our method.
Our framework consists of two branches: a DL classification branch (e.g., ResNet152~\cite{he2016deep}) and a SAM-based feature extraction branch.
On the SAM-based feature extraction branch, a frozen SAM image encoder is used to extract features from different blocks (Section~\ref{sam-image-encoder}), which are passed to SLCAs (Section~\ref{SLCA}) to compute the attention weights.
The computed attention weights are integrated into the corresponding stages of the DL classification branch via a residual connection to help the model focus on spatially relevant local regions in the image (Section~\ref{model}).

\subsection{SAM Image Encoder: A Frozen Feature Extractor} \label{sam-image-encoder}
The SAM image encoder, a Vision Transformer (ViT)~\cite{dosovitskiy2020image}, is pre-trained with Masked Auto-Encoders~\cite{he2022masked} using over 1 billion tasks in 11 million images.
It accepts an input image of any size and reshapes it to size $1024 \times 1024$. 
Then the image is converted to sequential patch embeddings
with patch size $16 \times 16$ and embedding size 256.
Through 32 Transformer blocks with window attention and a Conv block, the output of the image encoder is of size $64 \times 64 \times 256$, where 256 is the channel number. The outputs of patch embedding, Transformer blocks, and Conv block are all of size $64 \times 64 \times 256$.

In this paper, to efficiently leverage the SAM image encoder as a feature extractor to extract segmentation-based features of the input image:
(I) The weights of the SAM image encoder are frozen during the training stage to reduce computational overhead.
(II) There are three different scales of pre-trained image encoders: ViT-h, ViT-l, and ViT-b. We use ViT-h in all our experiments, considering the trade-off between real-time performance and accuracy.
By doing so, we ensure that SAM's pre-trained segmentation capabilities are preserved without the need for additional fine-tuning.
%

\subsection{Spatially Localized Channel Attention (SLCA)}\label{SLCA}
The features extracted by the SAM image encoder contain important spatial information of the image (see Fig.~\ref{fig:sam-cam}). It is crucial to utilize these features to enable the DL classification model to focus on spatially relevant regions in the image~\cite{mishra2022data,zhang2024ihcsurv,liang2020intracker,wang2024path}. 
Hence, a pertinent question is: How to effectively utilize the features obtained by the SAM image encoder to enable the DL classification model to focus on spatially meaningful local regions?

To address this question, we propose a new attention mechanism called Spatially Localized Channel Attention (SLCA), which enables the model to focus on what is spatially meaningful at a local level.
Specifically, we aggregate local spatial features of the feature map using spatial local average pooling, as shown in Fig.~\ref{fig:framework}(b). 
The locally spatially aggregated features are then passed through two Conv blocks and a sigmoid activation function to attain the spatially localized channel attention.
To reduce the number of parameters, we first reduce the channels of the locally spatially aggregated features by a factor of $r$ to obtain reduced features, and then increase the channels back to $C$.
Finally, the locally spatially aggregated features are passed through the sigmoid activation function to achieve the spatially localized channel attention.
Given a feature map $F$, the spatially localized channel attention weights can be computed as: 
\begin{equation}
    \text{SLCA}(F) = \sigma(Conv_2(ReLU(Conv_1(SLAP(F))))),
\end{equation}
where $SLAP$ is spatial local average pooling, $Conv_1$ and $Conv_2$ are $1\times 1$ Conv blocks, ReLU is the rectified linear unit activation function, and $\sigma$ is sigmoid activation function.

By assigning higher attention weights to channels that correspond to important spatial features, our SLCA focuses on what is spatially meaningful in an intermediate feature map (see Fig.~\ref{fig:vis}).
Importantly, SLCA is computationally efficient, adding minimal overhead while providing significant improvement in classification performance (see Table~\ref{tab:fusion-meth}).
We note that ECA-Net~\cite{wang2020eca} focuses on learning effective channel attention more efficiently, while our SCLA aims to capture the locally (spatially-local) meaningful aspects of channel attention.

\subsection{Adapting SAM Image Encoder for Medical Image Classification} \label{model}
As shown in Fig.~\ref{fig:framework}(a), we extract features from the SAM image encoder’s patch embedding, the 1st, 16th, and 32nd Transformer blocks, and the Conv block.
These features are passed to SLCA to compute their attention weights, which are then incorporated into the corresponding stages of the DL classification branch via residual connections.
These attention weights guide the model to focus on locally meaningful intermediate features, ensuring the classification model to prioritize relevant regions in its decision-making process and enhancing its performance.

To better utilize segmentation-based features that capture important spatial and contextual details of the image, we append the SAM image encoder with an additional task-specific projector head for fine-tuning.
As illustrated in Fig.~\ref{fig:framework}(c), we first resize the input feature map to size $56 \times 56$ using interpolation.
The resized feature map is then passed through a Conv block, consisting of a $3 \times 3$ kernel, batch normalization (BN), and a rectified linear unit (ReLU) layer, to increase the number of channels to $C$, where $C$ (e.g., 1024 or 2048) is the number of output feature channels from the last stage of the DL classification branch.
Finally, three Conv blocks, each consisting of a $3 \times 3$ kernel, stride of 2, BN, and ReLU, are used to downsample the feature map from a $56 \times 56$ size to $7 \times 7$.
The output of the projector head is concatenated with the DL classification branch’s features for further refinement, and the fused features are passed to a classifier to make the final prediction.

\section{Experiments and Results} \label{exp}
\label{sec:exp}
\textbf{Datasets.}
(1) RetinaMNIST: This dataset~\cite{yang2021medmnist,dataset20202nd} contains 1600 retina fundus images, including 1080 training, 120 validation, and 400 test images. Following~\cite{yang2021medmnist}, our experiments are focused on ordinal regression (5 classes).
(2) BreastMNIST: The dataset~\cite{yang2021medmnist,al2020dataset} contains 780 breast ultrasound images, including 546 training, 78 validation, and 156 test images. 
Following~\cite{yang2021medmnist}, we simplify the task as binary classification by combing normal
and benign as positive and classifying them against malignant as negative.
(3) The ISIC 2017 skin lesion classification dataset (ISIC 2017): The dataset~\cite{codella2018skin} contains 2000 training, 150 validation, and 600 test images. Our experiments are focused on task-3A: melanoma detection, following SAMAug-C~\cite{gu2024boosting}.

We resize all the images of each dataset to size 224$\times$224. Our experiments conduct 5 runs using different seeds, and present the average outcomes of the experimental results.

\textbf{Implementation Details.}
We use PyTorch for the experiments. The model is trained on an NVIDIA Tesla V100 Graphics Card (32 GB GPU memory) using the AdamW optimizer~\cite{loshchilov2017decoupled} with a weight decay = $0.005$. 
For all experiments, the learning rate is set to 0.0001. The number of training epochs is 100 for the RetinaMNIST and BreastMNIST datasets, and 400 for the ISIC 2017 dataset. The batch size in each case is set to the maximum size allowed by the GPU.
Standard data augmentation (e.g., random flip, crop, etc.) is applied to avoid overfitting.

\begin{table*}[ht!]
    \centering
     \caption{Results of using different DL classification models on the RetinaMNIST and BreastMNIST datasets.}\label{tab:res_main}
    \resizebox{1.0\textwidth}{!}{
    \begin{tabular}{lcccccccccccc}
        \toprule
        \multirow{3}{*}{\textbf{Method}} & \multicolumn{6}{c}{RetinaMNIST} & \multicolumn{6}{c}{BreastMNIST} \\
        \cmidrule(lr){2-7} \cmidrule(lr){8-13}
        {} & \multicolumn{2}{c}{10\%} & \multicolumn{2}{c}{50\%} & \multicolumn{2}{c}{100\%} & \multicolumn{2}{c}{10\%} & \multicolumn{2}{c}{50\%} & \multicolumn{2}{c}{100\%} \\
        \cmidrule(lr){2-3} \cmidrule(lr){4-5} \cmidrule(lr){6-7}
        \cmidrule(lr){8-9} \cmidrule(lr){10-11} \cmidrule(lr){12-13}
        {} & Acc  & AUC & Acc & AUC & Acc & AUC & Acc & AUC & Acc & AUC & Acc & AUC \\
        \midrule\midrule
        ResNet152~\cite{he2016deep} &52.50  &75.28  &64.00  &81.74  & 65.50 & 84.71 &81.41  &73.52  &87.18  &88.38  &91.03  & 91.90 \\\midrule
              \begin{tabular}[c]{@{}l@{}} ResNet152\\ + SAMAug-C~\cite{gu2024boosting} \end{tabular} &54.75   &  74.84 &64.50   &82.60  &66.00  &86.30  &82.69   &  78.59 &87.18  &88.16   &91.74   &92.61   \\\midrule
         \begin{tabular}[c]{@{}l@{}} ResNet152\\ + SAM (Ours) \end{tabular}  &56.50  &76.31  &65.25  &83.87  & 67.42 & 86.05 & 85.33 &78.69  & 90.10 &88.87  &92.95  & 93.15 \\\midrule\midrule
        SENet154~\cite{hu2018squeeze} &51.75  &73.94  &62.75  &80.46  & 66.33 &85.36 &83.33  &79.71  &88.46  &88.74  & 92.31 &92.82  \\\midrule
        \begin{tabular}[c]{@{}l@{}} SENet154\\ + SAMAug-C~\cite{gu2024boosting} \end{tabular} &52.25   & 76.19  &64.00   &83.40  &67.00  &86.82  & 83.69  &81.12   &89.74  & 89.60  &92.38   & 93.15  \\\midrule
        \begin{tabular}[c]{@{}l@{}} SENet154\\ + SAM (Ours) \end{tabular} &56.50  &76.55  &65.75  & 83.84 & 68.50 & 86.99 &86.54  & 81.62 &90.74  &90.65  & 94.87 &94.35  \\\midrule\midrule
        Swin Transformer v2~\cite{liu2022swin} &51.25  &72.53  &63.00  &82.65  & 68.25 & 86.18 &80.77  &77.86  &84.62  &82.21  &89.74  &91.40  \\\midrule
       \begin{tabular}[c]{@{}l@{}} Swin Transformer v2\\ + SAMAug-C~\cite{gu2024boosting} \end{tabular} &53.75   & 77.08  &63.25   & 84.44 &68.75&88.63   & 82.05 & 77.19  &84.33  &83.83   &89.46   &   91.77\\\midrule
         \begin{tabular}[c]{@{}l@{}}Swin Transformer v2\\ + SAM (Ours) \end{tabular}   &57.00  &79.64  &65.25  & 84.34 & 69.75 & 88.52 &83.54  &82.89  & 88.90 & 87.00 & 91.88 &93.47  \\
        \bottomrule
    \end{tabular}}
\end{table*}

\begin{table*}[ht!]
    \centering
     \caption{Results of using different DL classification models on the ISIC 2017 dataset.}\label{tab:res_main_isi17}
    \resizebox{0.7\textwidth}{!}{
    \begin{tabular}{lccccccc}
        \toprule
        \multirow{4}{*}{\textbf{Method}} & \multicolumn{6}{c}{ISIC 2017 } \\
        \cmidrule(lr){2-7} & \multicolumn{2}{c}{10\%} & \multicolumn{2}{c}{50\%} & \multicolumn{2}{c}{100\%} \\
        \cmidrule(lr){2-3} \cmidrule(lr){4-5} \cmidrule(lr){6-7}
        {} & Acc  & AUC & Acc & AUC & Acc & AUC  \\
        \midrule\midrule
        ResNet152~\cite{he2016deep} &74.00   & 70.12  & 80.67  &70.17  &82.00  & 74.38\\\midrule
              \begin{tabular}[c]{@{}l@{}} ResNet152\\ + SAMAug-C~\cite{gu2024boosting} \end{tabular} & 76.50  &  71.43 &81.17   & 71.24 &84.00  &75.91 \\\midrule
         \begin{tabular}[c]{@{}l@{}} ResNet152\\ + SAM (Ours) \end{tabular}  &79.00   &  73.22 & 83.17  & 78.56 &84.83  & 80.77\\\midrule\midrule
        SENet154~\cite{hu2018squeeze} & 75.33  &65.81   & 81.33  & 68.42 & 83.67 &  78.13 \\\midrule
        \begin{tabular}[c]{@{}l@{}} SENet154\\ + SAMAug-C~\cite{gu2024boosting} \end{tabular} & 76.17  & 69.50  &81.67   & 71.05 & 83.83 & 79.77  \\\midrule
        \begin{tabular}[c]{@{}l@{}} SENet154\\ + SAM (Ours) \end{tabular} & 76.67  & 71.13  & 82.83  &77.16  & 84.33 &  80.64\\\midrule\midrule
        Swin Transformer v2~\cite{liu2022swin} &76.33   & 68.44  & 80.00  &72.75  & 82.67 & 79.06 \\\midrule
       \begin{tabular}[c]{@{}l@{}} Swin Transformer v2\\ + SAMAug-C~\cite{gu2024boosting} \end{tabular} &77.33   & 68.96  &81.50   & 73.20 & 82.83 & 78.09  \\\midrule
         \begin{tabular}[c]{@{}l@{}}Swin Transformer v2\\ + SAM (Ours) \end{tabular}   & 77.17  & 71.92  & 82.17  &78.32  &83.50  & 79.85 \\
        \bottomrule
    \end{tabular}}
\end{table*}

\begin{table}[t]
    \centering
    \caption{Results of using different fusion methods and components on the RetinaMNIST dataset.}
    \label{tab:fusion-meth}
    \scalebox{1.0}{%
    \begin{tabular}{ll}
       \toprule
        Method &Acc  \\\midrule
        SENet154~\cite{hu2018squeeze}&	66.33	\\\midrule
         SENet154~\cite{hu2018squeeze} + add (no attention)&	60.25	\\\midrule
          SENet154~\cite{hu2018squeeze} + sigmoid&	66.75	\\\midrule
           SENet154~\cite{hu2018squeeze} + SLCA&	68.00	\\\midrule
            SENet154~\cite{hu2018squeeze} + SLCA + projector head&	68.50 	\\
      \bottomrule
    \end{tabular}
}
\end{table}

\begin{table}[t]
    \centering
     \caption{Results of using features from same block or different blocks of the SAM image encoder on the RetinaMNIST dataset.}
    \label{tab:position}
    \scalebox{1.0}{%
    \begin{tabular}{ll}
       \toprule
        Method &Acc  \\\midrule
        Swin Transformer v2~\cite{liu2022swin}&	68.25	\\\midrule
         \begin{tabular}[c]{@{}l@{}}Swin Transformer v2~\cite{liu2022swin} w/ all five features\\ from patch embedding (PE)\end{tabular}&	66.25	\\\midrule
         \begin{tabular}[c]{@{}l@{}}Swin Transformer v2~\cite{liu2022swin} w/ all five features\\ from the 1st Transformer block\end{tabular}&	67.50	\\\midrule
          \begin{tabular}[c]{@{}l@{}}Swin Transformer v2~\cite{liu2022swin} w/ all five features\\ from the 16th Transformer block\end{tabular}&	67.75	\\\midrule
           \begin{tabular}[c]{@{}l@{}}Swin Transformer v2~\cite{liu2022swin} w/ all five features\\ from the 32nd Transformer block\end{tabular}&	65.75	\\\midrule
           \begin{tabular}[c]{@{}l@{}}Swin Transformer v2~\cite{liu2022swin} w/ all five features\\ from the Conv block\end{tabular}&	66.75	\\\midrule
            \begin{tabular}[c]{@{}l@{}} Swin Transformer v2~\cite{liu2022swin} w/ features \\from PE, the 1st, 16th, and 32nd Transformer blocks,\\ and the Conv block (ours) \end{tabular}&	69.75	\\
      \bottomrule
    \end{tabular}
}
\end{table}

\begin{figure}[t]
    \centering
    \includegraphics[width=1.0\linewidth]{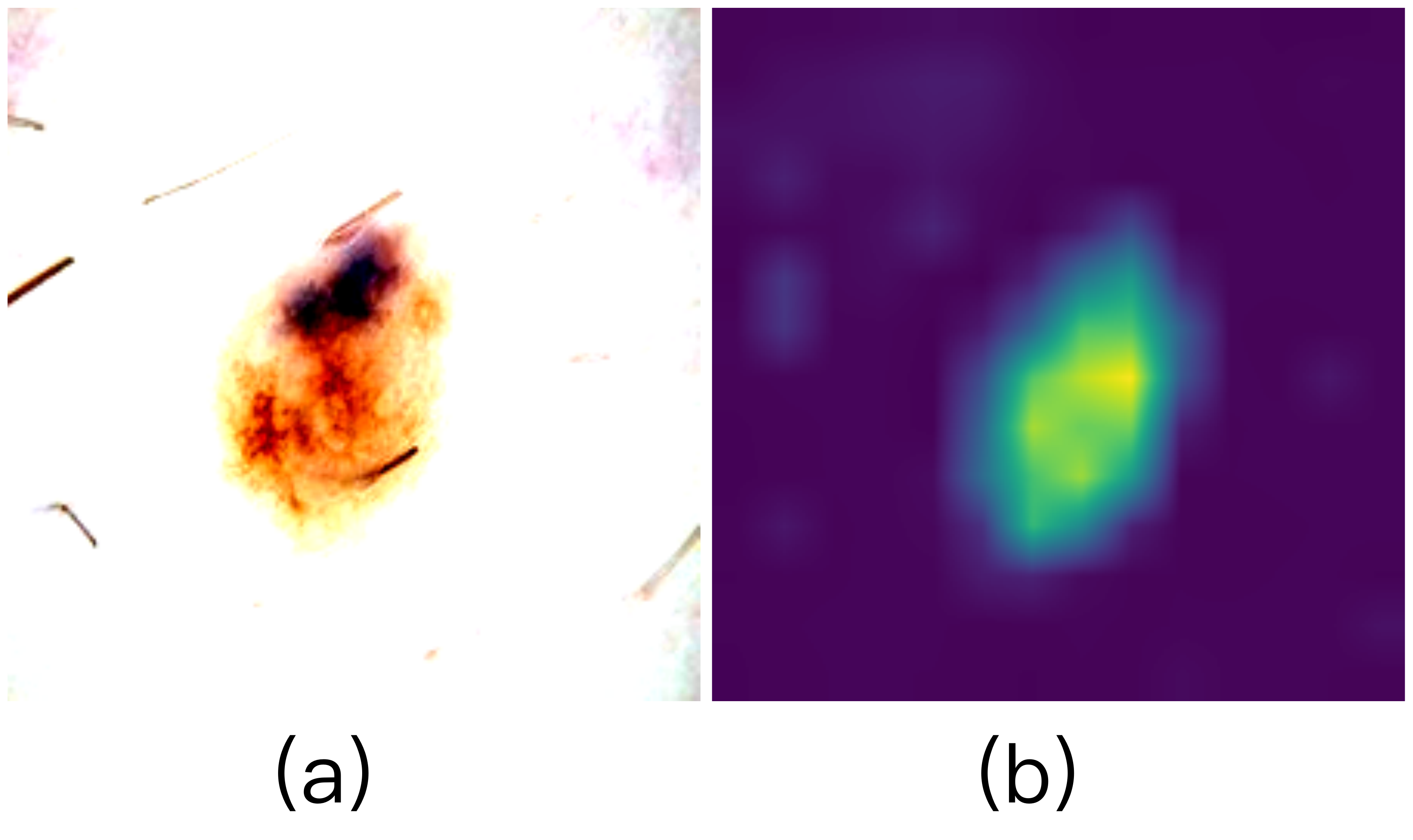}
    \caption{Visual result demonstrating that meaningful semantics are extracted by the SAM Image Encoder on the ISIC 2017 dataset: (a) the input image, and (b) the heat map of feature extracted by the SAM Image Encoder.}
    \label{fig:sam-cam}
\end{figure}

\textbf{Experimental Results.}
To evaluate our approach, we use two prominent CNN-based models (ResNet152~\cite{he2016deep} and SENet154~\cite{hu2018squeeze}), and one of the latest Transformer-based models (Swin Transformer v2~\cite{liu2022swin}).
Specifically, we incorporate the features extracted from the SAM image encoder into the corresponding stages of these DL classification models, as shown in Fig.~\ref{fig:framework}(a).
Tables~\ref{tab:res_main} and~\ref{tab:res_main_isi17} present the experimental results on the three datasets, from which one can observe the following.
(I) When incorporating the features extracted from the SAM image encoder into these DL classification models, the performance using 100\% of the training data is considerably improved. This can be attributed to the segmentation-based features from the SAM image encoder, which capture important spatial and contextual details of the images.
(II) The less the annotations used, the greater the performance improvements. This is a desirable advantage to tasks with many biomedical objects and very limited labels. In particular, when using only 10\% of the labeled training data, the performance is improved by 5.75\% in accuracy for Swin Transformer v2~\cite{liu2022swin} on RetinaMNIST and by 5.0\% in accuracy for ResNet152~\cite{he2016deep} on ISIC 2017. This demonstrates the data-efficiency of our method.
(III) Our method outperforms SAMAug-C~\cite{gu2024boosting} on all three datasets across almost all metrics with three different percentages of training data. This validates that our method effectively leverages SAM's capability to capture spatially localized features, providing valuable segmentation-based information for improving medical image classification (see Fig.~\ref{fig:sam-cam}).

\textbf{Ablation Study.}
We conduct ablation experiments to examine the contributions of key individual components in our approach, as shown in
Table~\ref{tab:fusion-meth}. One can see the following. (I) There is a 1.67\% improvement in accuracy after using SLCA to compute the attention weights, which shows the effectiveness of SLCA. (2) There is a further 0.5\% improvement in accuracy after adding the projector head. This indicates the benefit of the projector head in further leveraging the segmentation-based features from the SAM image encoder.

\begin{figure}[t]
    \centering
    \includegraphics[width=1.0\linewidth]{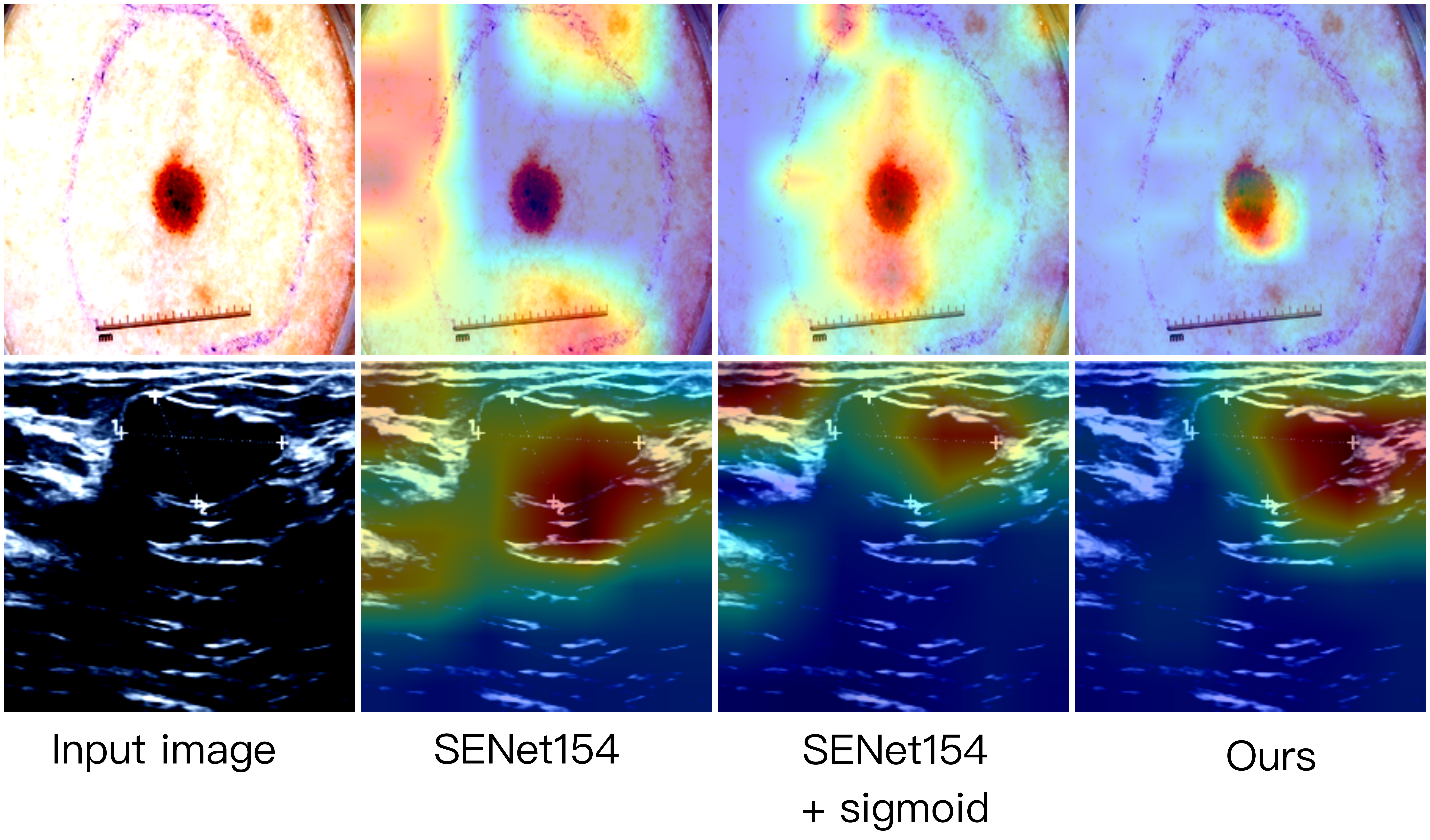}
    \caption{Visual examples of different attention methods on the ISIC 2017 (top row) and BreastMNIST (bottom row) datasets.}
    \label{fig:vis}
\end{figure}

\textbf{Effects of Fusion Choices.}
We conduct experiments to investigate effects of different fusion methods. We consider the following fusion methods: (1) Directly adding the features from the SAM image encoder to the corresponding stages of DL classification models without using an attention mechanism; (2) using a sigmoid (instead of SLCA) to compute attention weights, which are then applied via a residual connection.
As shown in Table~\ref{tab:fusion-meth}, directly adding features from the SAM image encoder to the corresponding stages of DL classification models leads to a decline in performance, while using a sigmoid to compute attention weights results in a minor improvement. However, using SLCA results in a larger improvement. This validates the effectiveness of our SLCA.
Fig.~\ref{fig:vis} shows that our method is capable to enhance the models' focus on spatially relevant regions in images.

\textbf{Effects of Block Choices of SAM Image Encoder for Extracting Features.}
We extract features from different blocks of the SAM image encoder.
It is important to investigate whether features for different stages of the DL classification model should be extracted from the same block.
Specifically, we extract five features from each of the following blocks: the patch embedding, the 1st, 16th, and 32nd Transformer blocks, and the Conv block.
As shown in Table~\ref{tab:position}, extracting all five features from the same block results in a performance drop, while extracting features from the embedding, 1st, 16th, and 32nd Transformer blocks, and the Conv block results in a performance improvement. This validates the effectiveness of our chosen blocks for feature extraction.

\section{Conclusions} \label{concl}
In this paper, we presented a new framework for adapting SAM effectively for medical image classification.
We first leveraged the SAM image encoder as a frozen feature extractor to capture segmentation-based features that contain important spatial and contextual details of the images.
Then, we proposed a novel Spatially Localized Channel Attention (SLCA) mechanism, which allows DL classification models to focus on what is spatially meaningful in the images and feature maps.
Experiments on three public datasets demonstrated the efficacy of our new approach.

\section{Acknowledgement}
\label{sec:acknowledgements}
This research was supported in part by the Graduate Assistance in Areas of National Need (GAANN) grant.


\bibliographystyle{IEEEtran} 

\bibliography{ref}
\end{document}